# Optimizing Language Models for Grammatical Acceptability: A Comparative Study of Fine-Tuning Techniques

Farley Knight
Georgia Tech
fknight9@gatech.edu

Ghada Jerfel
Georgia Tech
gjerfel3@gatech.edu

Sze Chung Ho
Georgia Tech
sho64@gatech.edu

Shobhit Ratan
Georgia Tech
sratan6@gatech.edu

**Abstract**

This study explores the fine-tuning (FT) of the Open Pre-trained Transformer (OPT-125M) for grammatical acceptability tasks using the CoLA dataset. By comparing Vanilla-Fine-Tuning (VFT), Pattern-Based-Fine-Tuning (PBFT), and Parameter-Efficient Fine-Tuning techniques (PEFT) like Low-Rank Adaptation (LoRA), we demonstrate significant improvements in computational efficiency while maintaining high accuracy. Our experiments reveal that while VFT achieves the highest accuracy (81.2%), LoRA enhancing FT by reducing memory usage and iteration time by more than 50%, and increases accuracy in PBFT case. Context Distillation (CD), though computationally efficient, underperformed with accuracy around 31%. Our findings contribute to democratizing access to large language models (LLM) by reducing computational barriers.

# Introduction

## Objective

The primary objective of this project is to adapt the OPT language model to determine grammatical correctness in sentences. By leveraging FT methods, the goal is to enable the model to learn from labeled examples and patterns indicative of grammaticality. This adaptation must balance accuracy and computational efficiency to ensure practical applicability.

## Background on Fine-Tuning (FT)

Fine-tuning has become a key strategy in machine learning, adapting pre-trained models (PTM) for specific tasks. Building on the advances by Radford et al. (2019), which showcased the few-shot learning abilities of language models, FT refines these models for specific datasets or domains. This approach utilizes the embedded knowledge in PTM, speeding up development, lowering costs, and improving domain-specific applications (DSA).

## Fine-Tuning Techniques

This project explores several FT techniques to optimize performance and efficiency:

- **VFT** updates all model parameters .
- **PEFT** like LoRA updates only a subset of parameters.
- **PBFT** uses structured prompts for models behavior.
- **CD** uses teacher-student transfer knowledge .

## Categories of FT

Recent advancements in FT techniques that we used are:

- **Supervised FT:** Adapts models using labeled datasets, achieving precision for task-specific outputs. However, it relies heavily on high-quality labeled data, which can be resource-intensive .
- **Instruction-Based FT:** Uses task-specific prompts thus enhancing generalizability and cost efficiency .

## Significance and Applications

If successful, the proposed methods will significantly create helpful, honest, and harmless model . Such a model would be highly general, but we focus on grammatical correctness task and useful word editor. By reducing computational costs, making it accessible to organizations with limited resources while reducing the environmental impact of model adaptation.

## Dataset Description and Preparation.

This project uses the Corpus of Linguistic Acceptability (CoLA), comprises 10,657 English sentences annotated for acceptability classification tasks. . To ensure robustness, the dataset is split into in-domain (ID;training and validation) and out-of-domain (OOD) test sets. Sentences are tokenized to fit model constraints, and the dataset is balanced to mitigate class bias.

# Approach

This research focuses on adapting Facebook's OPT-125M model for binary classification using CoLA to assess sentence grammaticality. Due to time constraints, we prioritized developing the 125M model. We initially used grid search for hyperparameter optimization

(HPO) of batch size, learning rate, and optimizer selection, within a few-shot learning framework. In this setting, the model is trained to use only small number of labeled examples. However, as shown by Bergstra and Bengio , grid search could be more efficient for high-dimensional parameter spaces. To improve this, we adopted Bayesian optimization via Hyperopt , utilizing Tree-structured Parzen Estimators (TPE) to optimize efficiently for the OPT models and CoLA dataset requirements.

The Hugging Face Transformers library was the foundation of our implementation, which involved modifying the training process for task-specific tokenization, FT parameters, and optimizer flexibility. We also enhanced logging and metrics evaluation to assess ID and OOD performance comprehensively. Our methodology aligns with Liu et al. findings in RoBERTa, indicating that targeted adjustments to FT pipelines can improve downstream task performance. Additionally, the Hugging Face documentation was crucial for configuring training arguments and evaluation metrics.

## Vanilla Fine Tuning

VFT involves adapting a pre-trained model to a specific downstream task by training it on task-specific labeled data. This approach aligns with foundational methodologies outlined by Howard and Ruder in their work on Universal Language Model FT (ULMFiT), emphasizing task-specific adaptation's effectiveness for improving downstream performance. We utilized this approach for the OPT-125M model by adding a classification head and retraining the model on the CoLA dataset, using cross-entropy loss to optimize for binary classification. Additionally, inspired by Devlin et al. in the BERT paper, we evaluated different optimizer choices such as Adam, AdamW, and SGD, ultimately selecting AdamW for its effectiveness in handling sparse gradients and compatibility with Transformer-based architectures. This methodology leverages the pre-trained model's ability to generalize across diverse NLP tasks, allowing for faster convergence and better performance on the specific task.

Our approach utilized BO with a custom training wrapper to streamline FT, efficiently identifying optimal hyperparameters and configurations while reducing computational overhead. This methodology aligns with Sun et al. , who emphasized the importance of systematic hyperparameter tuning for Transformer-based models.

## Pattern-Based Fine-Tuning

PBFT was developed as an extension of traditional FT approaches, incorporating structured prompting techniques to guide model behavior.

This methodology builds upon the work of  in prompt engineering strategies: minimal (appending "?"), GPT-3 style (explicitly asking about grammatical correctness), and eval-harness (using structured format with clear delineation). We utilized the OPT-125M model with a binary classification head, adapting these prompting patterns through a custom **COLADataPreparator** class that systematically transforms input sentences using template-specific lambda functions.

Our implementation leverages BO through the **HyperparameterOptimizer** class, which systematically explores both template selection and model configuration space. The optimization framework integrates with a custom **ModelBuilder** to handle dropout rates and architecture-specific parameters, while maintaining consistent tokenization across prompt variations. This approach allows for dynamic template selection during hyperparameter search, with the option to either fix the prompt pattern or include it as part of the optimization space alongside traditional parameters such as learning rate, batch size, and warmup ratio. The methodology emphasizes reproducibility through systematic logging (via a custom **MetricsTracker**), enabling detailed analysis of the interaction between prompting strategies and model performance.

## Low-Rank Adaptation

While LoRA has been applied by various researchers and practitioners, it is still considered relatively new since its introduction in 2021 in the paper "LoRA: Low-Rank Adaptation of Large Language Models" by Edward Hu et al. from Microsoft Research.

Over the past two years, the implementation of LoRA has become more widespread due to its low resource consumption during FT. This makes it an attractive option for adapting large models with limited computational resources.

We were interested in LoRA's reduction of the number of parameter updates during back-propagation. This would also reduce the over-fitting risk that few-shots can introduce considering the smaller sample sizes.

In our approach, we integrated LoRA within the pre-existing framework of VFT and PBFT by adding it's specific configuration to these models and not entirely modify them or their pre-trained weights matrices used for self-attention computations. Instead of FT all the parameters of the model, we only fine tune LoRA parameters.

## Context Distillation

CD employs a teacher-student architecture as described in , transferring knowledge from a deliberate reasoning process to an efficient inference model. Due to resource constraints, both models have the same parameters, but the teacher model incorporates a "scratchpad" mechanism for step-by-step reasoning, whereas the student model aims for more direct conclusions. Knowledge transfer via KL divergence loss minimizes output differences, aligned with the knowledge distillation approach from Hinton et al. (2015) . The teacher's parameters were frozen during training, and both models used the same initial instruction template ("Is this sentence grammatically correct?"). The teacher model included a scratchpad prompt ("Let me think about this step by step:") to foster improved reasoning. This design was based on the hypothesis that the student could learn to model the teacher's reasoning implicitly. We also expanded the hyperparameter space to include distillation-specific elements like temperature and distillation weight, managing the training process through hyperopt with TPE, similar to our PBFT implementation, but with additional parameters specific to distillation.

# Experiments and Results

## Success Criteria

Success was measured by accuracy, loss, memory usage, and epoch time. Accuracy served as the main metric for model performance on grammatical classification, while resource usage assessed efficiency. Criteria included achieving high accuracy on ID and OOD datasets and lowering resource consumption compared to VFT.

## Experiments

We conducted five experiments: VFT (baseline), VFT-LoRA, PBFT, PVFT-LoRA, and context distillation. Each experiment used the CoLA dataset, with ID and OOD splits to test generalization. LoRA reduced parameter updates while context distillation used a teacher-student framework.

## Quantitative and Qualitative Results

VFT achieved the highest accuracy but was costly, taking 189s per iteration with 4992MB memory. VFT-LoRA improved efficiency to 164s and 2898MB. PBFT-LoRA further reduced iteration time and memory usage from 6.9s and 6711MB to 3.7s and 2034MB. CD was the most efficient but had the lowest accuracy. PBFT-LoRA offers a good balance of accuracy and efficiency, while CD's performance indicates a need for refinement to enhance generalization.

| Method | Max In-Domain | Max Out-Domain |
|---|---|---|
| Vanilla | 0.8169 | 0.8120 |
| PBFT | 0.6512 | 0.6378 |
| Context Distillation | 0.3292 | 0.3106 |
| VFT-LoRA | 0.6960 | 0.6957 |
| PBFT-LoRA | 0.6692 | 0.6376 |

Table 1. Accuracy comparison of methods based on Max In-Domain and Max Out-Domain metrics.

| Method | Max Iteration Time (s) | Max Memory Usage (MB) |
|---|---|---|
| Vanilla | 189.4327 | 4991.8598 |
| PBFT | 6.9297 | 6711.4684 |
| Context Distillation | 0.4784 | 1791.4140 |
| VFT-LoRA | 164.8370 | 2898.5645 |
| PBFT-LoRA | 3.7354 | 2034.0820 |

Table 2. Efficiency comparison of methods based on Max Iteration Time and Max Memory Usage metrics.

## Why Did We Succeed?

PBFT-LoRA achieved a better balance between accuracy and efficiency compared to baseline PBFT, thanks to LoRA's parameter-efficient design, enabling effective model adaptation with fewer resources. VFT also excelled, reaching the highest accuracy (81.2%), particularly with the 125M model, showcasing its effectiveness when resources are ample.

## Why Did We Fail?

Despite its efficiency (0.48s/iteration and 1791MB memory), CD had the lowest accuracy (31.0%) due to the student model's limited capacity and difficulties in transferring grammatical knowledge from the teacher. Additionally, PBFT's high memory usage (6711MB) hindered practical usability, emphasizing the need for improvements in CD setups and PBFT resource optimization.

## Vanilla Fine-Tuning

The performance of the base model (OPT-125M) during FT is summarized in Figure 3, which provides a detailed breakdown of metrics across 20 epochs. The base model demonstrated steady improvements in ID and OOD accuracy over epochs. By the 20th epoch, the base model achieved an average ID accuracy of approximately 79.9% and an OOD accuracy of 78.2%, with a minimal difference of 1.7% between these metrics. This indicates the base model's ability to generalize reasonably well across distributions. However, the training process incurred higher memory usage, peaking at around 3100 MB, and required longer iteration times, particularly in later epochs. These observations underscore the need for optimization to enhance both performance and efficiency.

The search space was tailored to include key parameters influencing model performance, such as the number of epochs, batch size, learning rate, dropout probabilities, and optimizer choice. For instance, including num_epochs ranging from 2 to 50 allowed us to balance computational resources and convergence. At the same time, a log-uniform distribution for the learning rate ensured a fine-grained exploration of values spanning several magnitudes. Similarly, dropout probabilities were constrained between 0.0001 and 0.3 to optimize regularization without inducing excessive overfitting or underfitting.

The best hyperparameters identified through this process are reflected in the performance of the optimized model. As shown in Figure 1, the best model's average loss rapidly decreased during the initial epochs, stabilizing near 0.20 by epoch 50. This demonstrates the effectiveness of optimized hyperparameters—such as a learning rate of approximately $2.41 \times 10^{-5}$, a batch size of 128, and dropout probabilities of 0.26 (hidden) and 0.14 (attention)—in facilitating faster convergence. The best optimizer was identified as Adam, and the max sequence length of 256 ensured sufficient context while maintaining computational efficiency.

Overall, the comparison between the base and best models highlights the critical role of HPO in achieving superior performance and efficiency. By systematically exploring the hyperparameter space, we identified a configuration that improved accuracy and reduced computational overhead. The results underscore the importance of a well-planned fine-tuning approach for adapting large-scale pre-trained models to specific downstream tasks.

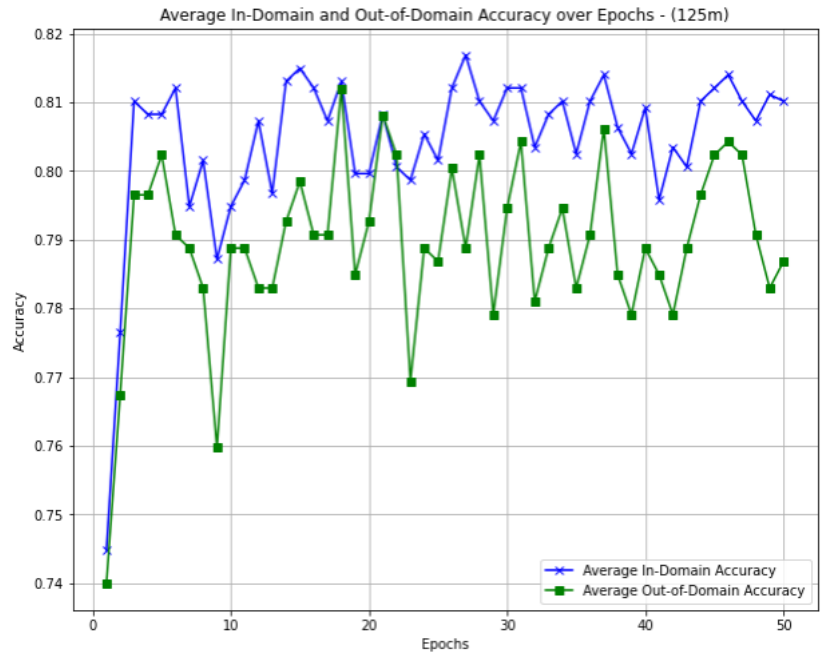

*Average Accuracy Iterations - 125m - COLA*

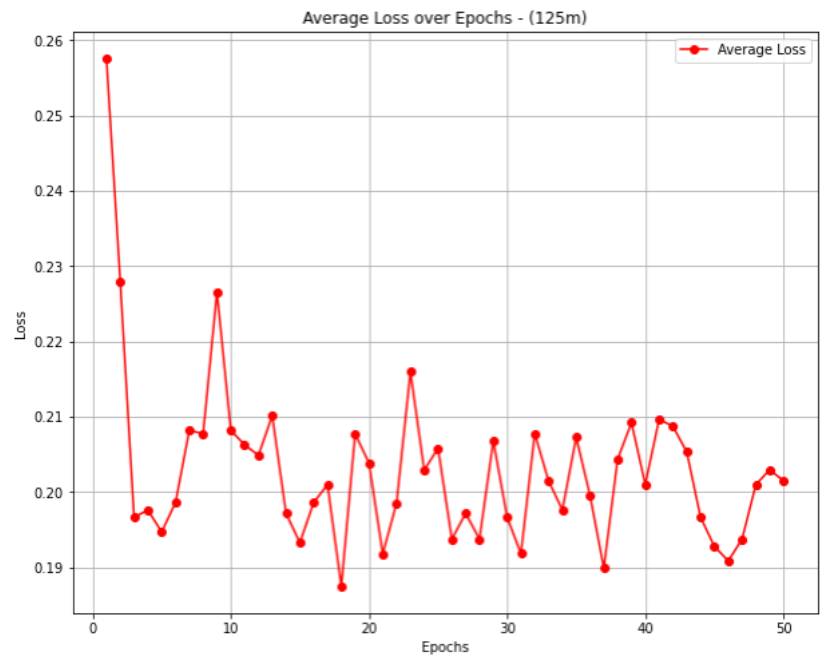

*Loss - 125m - COLA*

## Pattern-Based Fine-Tuning

The PBFT approach employs a comprehensive HPO strategy across multiple prompt patterns for 50 epochs each. The system uses the hyperopt framework to explore a parameter space with learning rates from $10^{-6}$ to $10^{-4}$, batch sizes from 2 to 16, and dropout rates up to 0.5. The optimization process evaluates ID and OOD accuracy with warmup ratios between 0 and 0.2 of total training steps. Dynamic sample sizes range from 2 to 32 examples per class, and epoch counts are optimized between 5 and 20 for efficient learning from limited examples.

Key findings from the optimization include that weight decay values near zero were most effective, increasing few-shot samples (up to 128), improved performance and larger batch sizes (up to 32) positively correlated with accuracy. The GPT-3 template slightly outperformed others (2-3% improvement).

Training dynamics revealed initial losses of $2.64 \times 10^{-6}$ with early fluctuations before stabilizing. ID accuracy peaked at 63.7% around epoch 26, while OOD accuracy reached 63.8% near epoch 32, indicating a strong generalization ability. The consistent gap between ID and OOD performance, averaging 2-4%, with a minimum difference of 1.3% at epoch 32, underscores the model's robustness. Training efficiency also improved, with average iteration times decreasing from 6.9 seconds to 5.3 seconds and memory usage falling from 6.7GB to 2.3GB, indicating effective resource management.

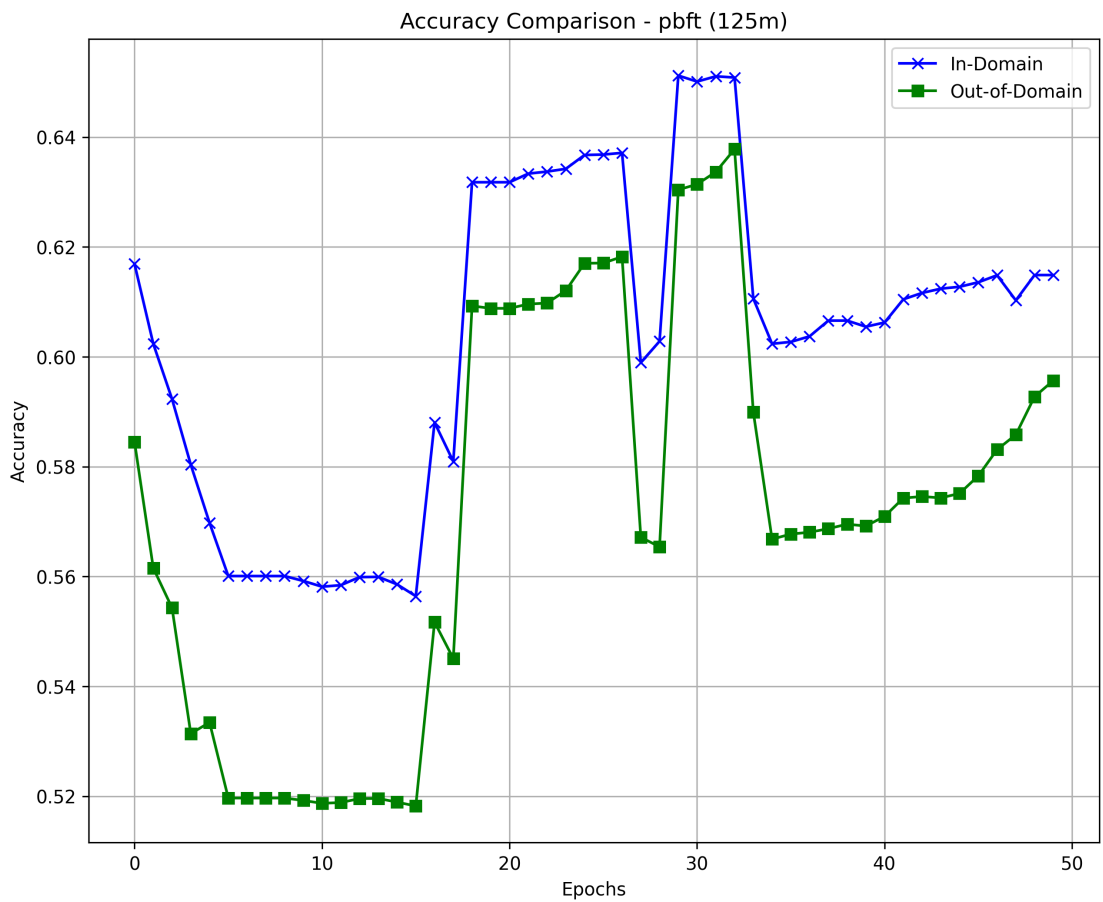

*Accuracy Comparisons- PBFT - 125m - COLA*

As other methods, the PBFT approach utilizes a comprehensive HPO strategy across multiple prompt patterns, for 50 epochs each. Through the hyperopt framework, the system explores a parameter space including learning rates from $10^{-6}$ to $10^{-4}$, batch sizes ranging from 2 to 16, and dropout rates up to 0.5. The optimization process evaluates both ID and OOD accuracy, with warmup ratios varying between 0 and 0.2 of total training steps. The training process employs dynamic sample sizes from 2 to 32 examples per class, while epoch counts are optimized between 5 and 20, allowing for efficient exploration of the model's capacity to learn from limited examples.

Optimization revealed several key findings. Weight decay values approaching zero proved most effective, while increasing the number of few-shot samples (up to our maximum of 128) consistently improved performance. Larger batch sizes, up to our hardware limit of 32, showed positive correlation with accuracy improvements. Between the mentioned templates (minimal, GPT-3, and eval-harness), the GPT-3 template showed only slightly better performance (2-3% improvement) over alternatives.

The training dynamics revealed interesting patterns across the 50 epochs. Starting with high initial losses around $2.64 \times 10^{-6}$, the model showed significant early fluctuations before stabilizing. The ID accuracy peaked at approximately 63.7% around epoch 26, while

OOD accuracy reached its maximum of 63.8% near epoch 32, demonstrating the model's ability to generalize. The gap between ID and OOD performance remained relatively small, averaging around 2-4% throughout training, with the smallest difference of 1.3% observed at epoch 32. Notably, training efficiency improved over time, with average iteration times decreasing from 6.9 seconds in early epochs to 5.3 seconds in later stages, while memory usage showed a consistent linear decrease from 6.7GB to 2.3GB, suggesting effective resource management during the training process.

## Low-Rank Adaptation

As with other methods, the PBFT approach utilizes a comprehensive HPO strategy across multiple prompt patterns for 50 epochs each. Through the hyperopt framework, the system explores a parameter space including learning rates from $10^{-6}$ to $10^{-4}$, batch sizes ranging from 2 to 16, and dropout rates up to 0.5. The optimization process evaluates both ID and OOD accuracy, with warmup ratios varying between 0 and 0.2 of total training steps. The training process employs dynamic sample sizes from 2 to 32 examples per class, while epoch counts are optimized between 5 and 20, allowing efficient exploration of the model's capacity to learn from limited examples.

Optimization revealed several key findings. Weight decay values approaching zero proved most effective while increasing the number of few-shot samples (up to our maximum of 128) consistently improved performance. Up to our hardware limit of 32, larger batch sizes positively correlated with accuracy improvements. Between the mentioned templates (minimal, GPT-3, and eval-harness), the GPT-3 template showed only slightly better performance (2-3% improvement) over alternatives.

The training dynamics revealed interesting patterns across the 50 epochs. The model showed significant early fluctuations before stabilizing, starting with high initial losses around $2.64 \times 10^{-6}$. The ID accuracy peaked at approximately 63.7% around epoch 26, while OOD accuracy reached its maximum of 63.8% near epoch 32, demonstrating the model's ability to generalize. The gap between ID and OOD performance remained relatively small, averaging around 2-4% throughout training, with the slightest difference of 1.3% observed at epoch 32. Notably, training efficiency improved over time, with average iteration times decreasing from 6.9 seconds in early epochs to 5.3 seconds in later stages, while memory usage showed a consistent linear decrease from 6.7GB to 2.3GB, suggesting effective resource management during the training process.

Our hyperparameter findings for VFT and PBFT with LoRA showed that $rank = 16$, $alpha = 64$, and $dropout = 0.2$ gave the best ID and OD accuracies and losses. The scatter plot in Figure 4 displays the lowest loss for a rank of 16 on the x-axis, a color corresponding to a dropout of 0.2 on the color map on the right, and a size of the circle corresponding to an alpha value of 64. Adapting LoRA in the attention layers, an important architectural feature did not improve accuracy compared to standard VFT. However, it showed better ID accuracy than PBFT, though they both had similar OD accuracy. Both approaches demonstrated stable training dynamics, with accuracy and loss metrics typically converging within the first 20-30 epochs (figure 5,6). It is worth noting that the demonstrated instability between epochs 30-50 is within 0.03 deviation, which is expected considering the randomness induced in the experiment through set_seed function in every run. The accuracy underperformance of VFT-LoRA compared to the best VFT model could be because the OPT-125M model is relatively small compared to other larger models. Reducing the parameter search space using LoRA could result in suboptimal accuracy performance. However, LoRA's improvements could be more pronounced. In larger models, fine-tuning all parameters is computationally very expensive, thus underlining the utility of LoRA's rank constraints in achieving better efficiency.

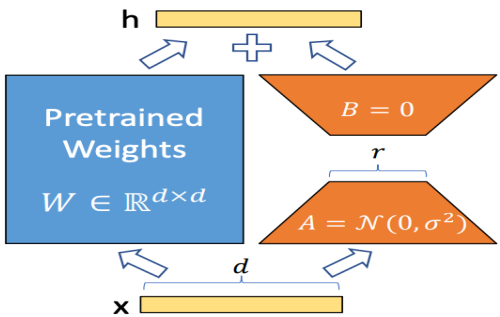

*Sample LoRA Architecture*

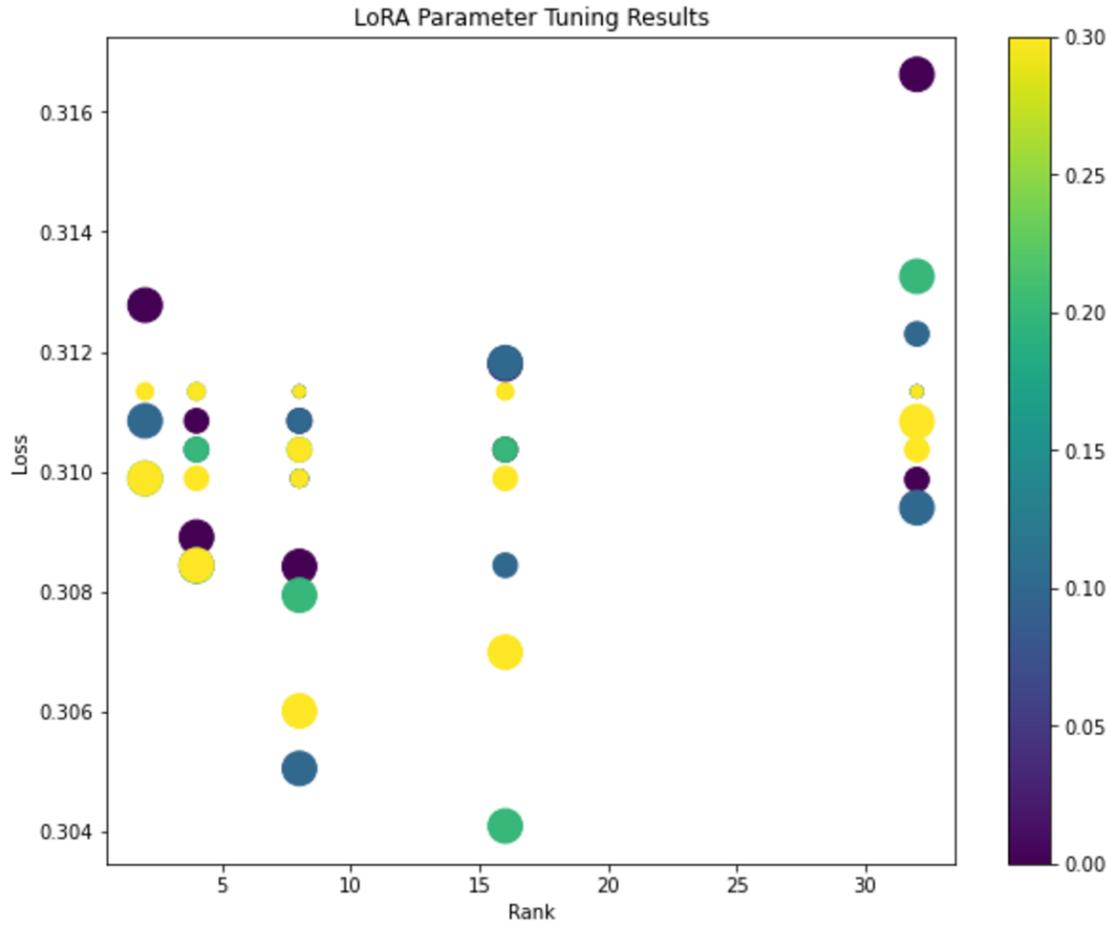

VFT LoRA Parameter Tuning Results

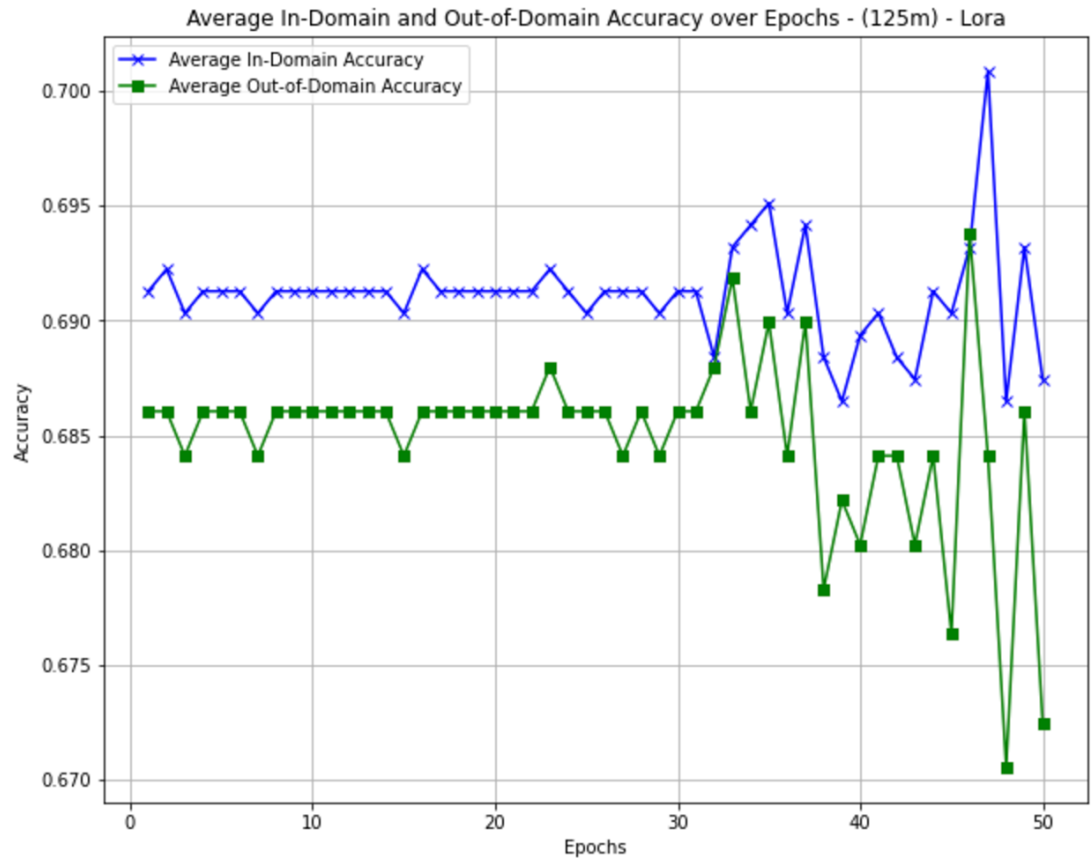

VFT LoRA Average In Out Domain Accuracy

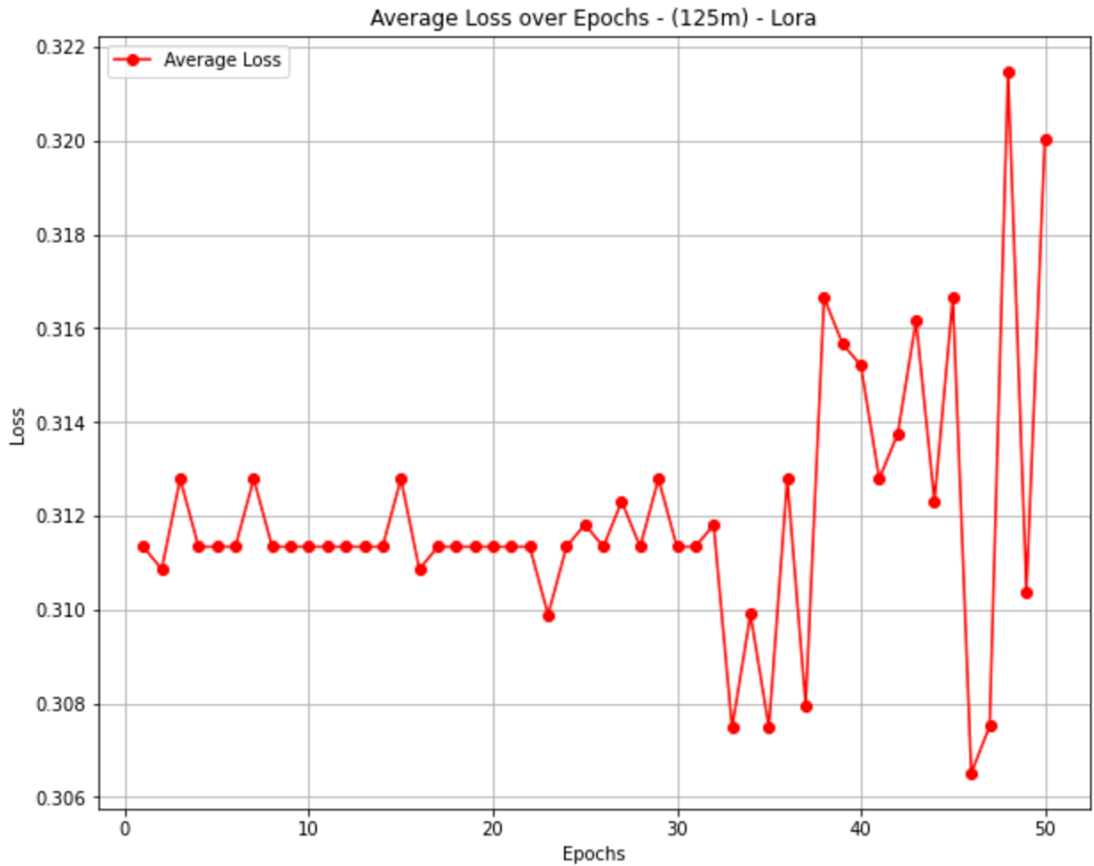

*VFT LoRA Average Loss Over Epochs*

## Context Distillation

CD, our novel approach, stood out from VFT and PBFT in several ways. During HPO, it demonstrated more effectiveness with lower dropout rates (around 0.1). A key hyperparameter was the temperature scaling factor for the KL divergence loss, where a value of 2.0 provided optimal stability. The distillation weight, which balances the contributions of cross-entropy and KL divergence, consistently converged to 0.5, indicating equal importance for teacher and student learning objectives.

CD, in contrast to other methods, displayed a unique accuracy behavior, showing more stability but generally lower accuracy (around 0.30-0.32). This stability, however, came at a performance cost, as higher initial accuracy scores might have resulted from overfitting due to the smaller sample size in our experiments.

The memory footprint for CD was stable throughout training, around 1.7-1.8GB, due to the efficient teacher-student architecture and lack of additional adaptation layers. Training times remained consistent across epochs, although the dual-model approach introduced some computational overhead.

Temperature scaling was vital for training stability, with values above 2.0 causing instability and below 1.0 hindering knowledge transfer. The distillation weight's consistent convergence to 0.5 across optimization runs suggests the equal importance of the teacher's guidance and the student's learning.

The training data over 50 epochs showed consistent behavior, with average loss fluctuating between 212.3 and 214.0, indicating early convergence. ID accuracy remained stable at 0.32-0.33, while OOD accuracy stayed around 0.30-0.31, maintaining a gap of about 0.015-0.022. The training process exhibited excellent stability regarding resources, with iteration times around 0.47 seconds and static memory usage near 1791MB, decreasing minimally to about 1786MB in the final epochs. This operational consistency aligns with previous memory footprint observations, highlighting the stability advantages of the context distillation approach.

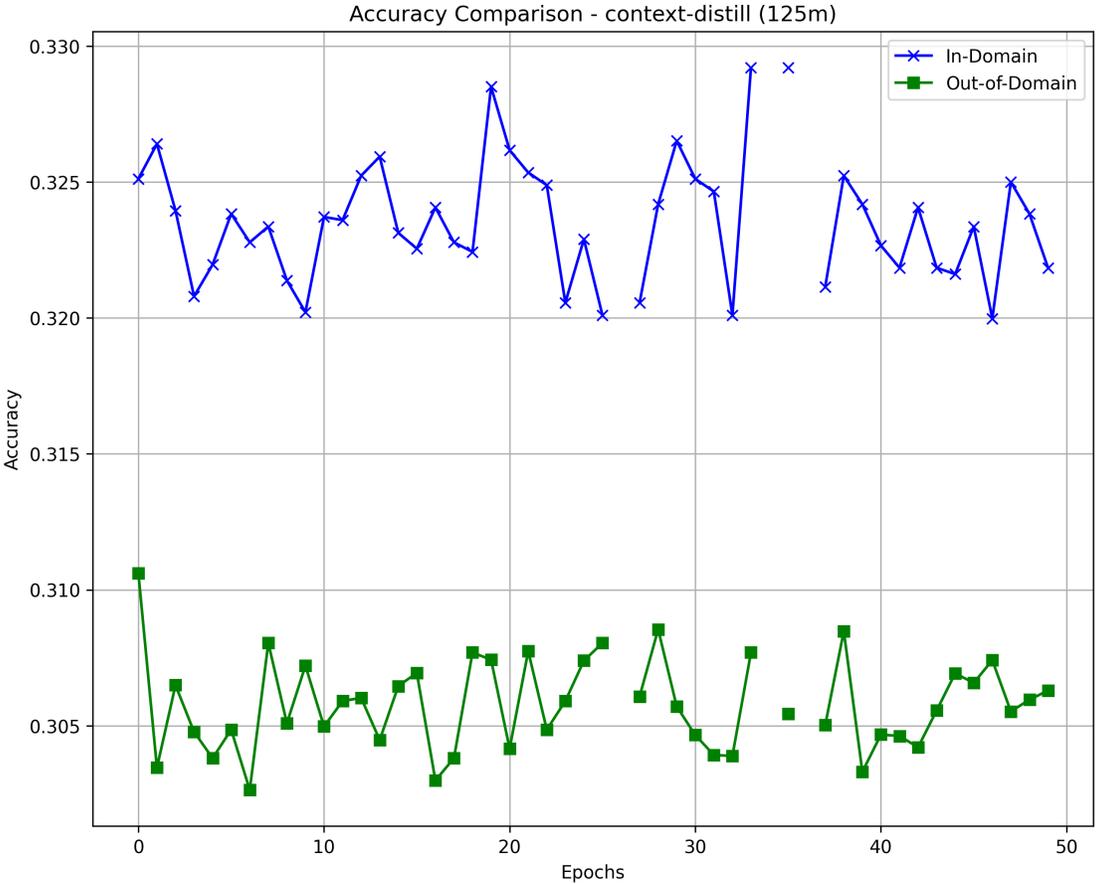

*Accuracy Comparisons- Context Distillation - 125m - COLA*

# Conclusion

This paper explored multiple fine-tuning approaches to adapt the OPT-125M model for grammatical correctness tasks using the CoLA dataset. The study demonstrated the potential of parameter-efficient methods, such as LoRA and PBFT-LoRA, to balance performance and computational efficiency.

VFT, while achieving the highest accuracy (81.2%), was computationally intensive, emphasizing the trade-off between performance and resource usage. Conversely, Context Distillation exhibited the lowest computational demands but underperformed in accuracy, underscoring the need for further refinement to enhance generalization capabilities.

Overall, the results validate the effectiveness of Bayesian Optimization for hyperparameter tuning and the practicality of parameter-efficient fine-tuning techniques. This work contributes to democratizing access to large language models by reducing computational barriers enabling their application across diverse domains such as grammar checking, education, and conversational AI. Future work may improve generalization in resource-efficient approaches, particularly for bigger data or models with more parameters.